\documentclass[10pt,twocolumn,letterpaper]{article}

\usepackage{cvpr}
\usepackage{times}
\usepackage{epsfig}
\usepackage{graphicx}
\usepackage{amsmath}
\usepackage{amssymb}
\usepackage{enumitem}
\usepackage[table,xcdraw,dvipsnames]{xcolor}
\usepackage{multirow}
\usepackage{tabularx}
\usepackage{graphicx}
\usepackage{makecell}

\usepackage[pagebackref=true,breaklinks=true,letterpaper=true,colorlinks,bookmarks=false]{hyperref}

\cvprfinalcopy %

\def\cvprPaperID{1226} %

\ifcvprfinal\fi
\begin{document}

\title{Domain-robust VQA with diverse datasets and methods but no target labels}

\author{Mingda Zhang \hspace{0.7cm} Tristan Maidment \hspace{0.7cm} Ahmad Diab \hspace{0.7cm} Adriana Kovashka \hspace{0.7cm} Rebecca Hwa\\
Department of Computer Science, University of Pittsburgh\\
{\tt\small \{mzhang, kovashka, hwa\}@cs.pitt.edu}
\hspace{1cm}
{\tt\small \{tdm51, ahd23\}@pitt.edu } 
\\
{\small \href{https://people.cs.pitt.edu/\%7Emzhang/domain_robust_vqa/}{\texttt{https://people.cs.pitt.edu/{\raise.17ex\hbox{$\scriptstyle\mathtt{\sim}$}}mzhang/domain{\char`\_}robust{\char`\_}vqa/}}}
}

\maketitle
\thispagestyle{empty}

\begin{abstract}
The observation that computer vision methods overfit to dataset specifics has inspired diverse attempts to make object recognition models robust to domain shifts. However, similar work on domain-robust visual question answering methods is very limited. Domain adaptation for VQA differs from adaptation for object recognition due to additional complexity: VQA models handle multimodal inputs, methods contain multiple steps with diverse modules resulting in complex optimization, and answer spaces in different datasets are vastly different. To tackle these challenges, %
we first quantify domain shifts between popular VQA datasets, in both visual and textual space. To disentangle shifts between datasets arising from different modalities, we also construct synthetic shifts in the image and question domains separately. Second, we test the robustness of different families of VQA methods (classic two-stream, transformer, and neuro-symbolic methods) to these shifts. Third, we test the applicability of existing domain adaptation methods and devise a new one to bridge VQA domain gaps, adjusted to specific VQA models. To emulate the setting of real-world generalization, we focus on unsupervised domain adaptation and the open-ended classification task formulation. 

\end{abstract}

\vspace{-0.5em}
\section{Introduction}
\label{sec:intro}
\vspace{-0.5em}

Visual question answering (VQA) borders on AI-completeness: it requires perception (visual and linguistic) and cognition. Despite the strong performance of recent VQA methods, they fall short of generalization and true reasoning: they are known to suffer from dataset bias \cite{Goyal_2017_CVPR}, require domain-specific languages or domain-specific executable program annotations~\cite{johnson2017inferring,Mao2019NeuroSymbolic}, or must be trained separately for each new dataset.

\begin{figure}[t]
    \centering
    \includegraphics[width=\linewidth]{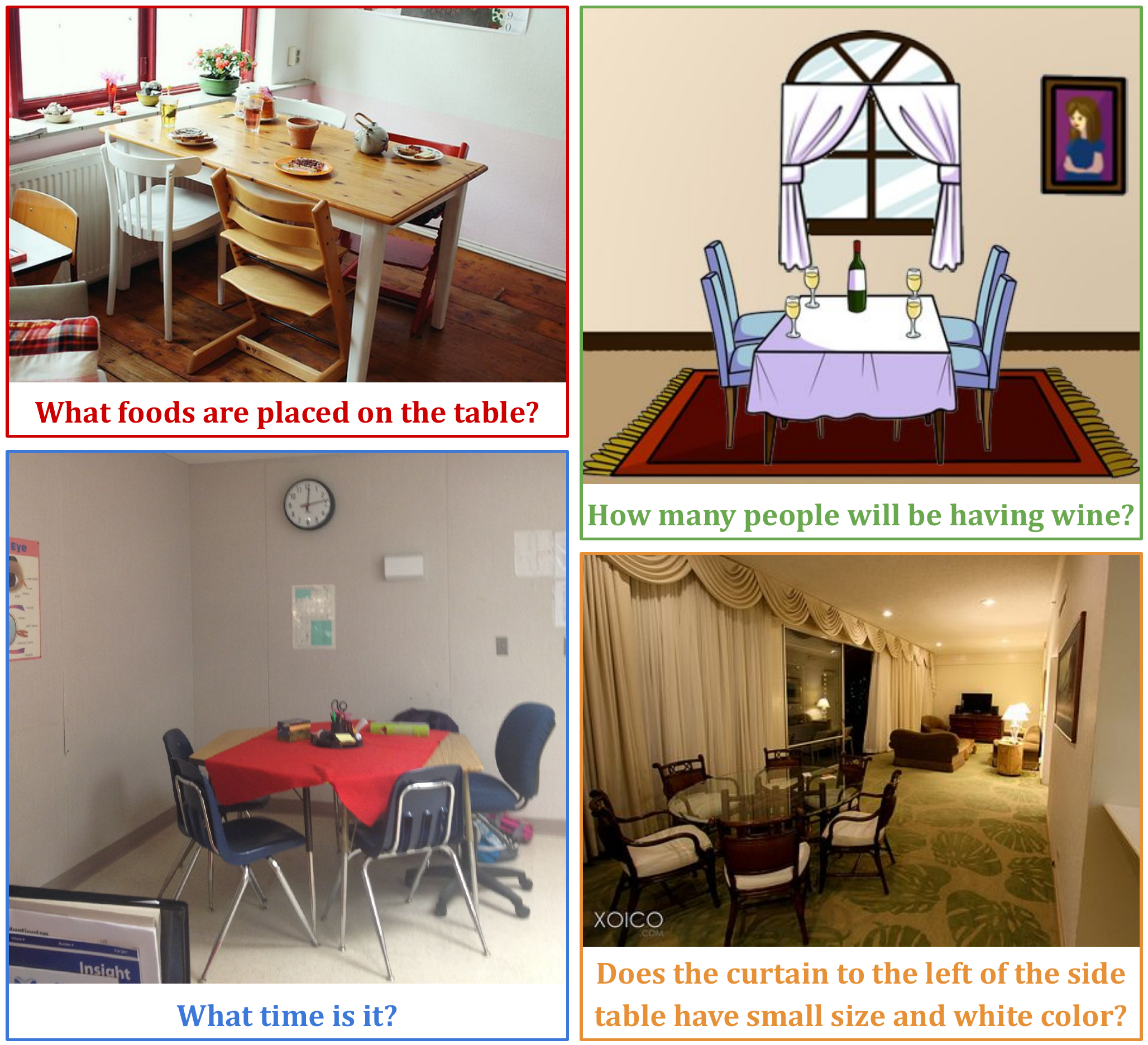}
    \caption{The same visual setting can be captured in different ways in VQA datasets, and paired with different information needs (questions). 
    They may require deduction using visual contents, reading from a specific region of the image, or reasoning about complex spatial relationships.
    All examples are selected from real VQA datasets, \ie \textcolor{red}{VQA v2}, \textcolor{ForestGreen}{VQA Abstract}, \textcolor{blue}{VizWiz} and \textcolor{orange}{GQA}.}
    \label{fig:concept}
\vspace{-1em}
\end{figure}

Prior work in domain adaptation for object recognition examines how robust methods are when trained and tested on different datasets (domains), and further proposes techniques to bridge domain gaps. In contrast, there is a shortage of analyses of how domain-robust visual question answering methods are. Importantly, domain adaptation techniques cannot successfully be applied in the VQA setting in a straight-forward manner. First, VQA models take \emph{inputs across multiple modalities}, each of which could contribute to the domain specificity of the trained models. Second, different VQA methods have \emph{multiple intermediate stages} and processing steps over the inputs, which makes optimization challenging. Domain adaptation techniques could be applied at multiple of these stages, and domain adaptation can be performed jointly or separately from VQA training, with varying success.
Third, \emph{answer spaces} in different datasets are vastly different. While domain adaptation methods exist to tackle non-identical answer spaces in object recognition, this setting is not very common. Conversely, in VQA, it is the norm, since many datasets are highly specialized (for example, VizWiz \cite{gurari2018vizwiz} contains special answers ``unanswerable'' or ``unsuitable image'' because image-question pairs are provided by visually impaired users). 

To tackle each of these challenges, we propose the following steps. First, to understand how the \emph{multiple modalities} contribute to domain shifts, we break down and measure both visual and textual domain shifts across datasets.
We disentangle shifts in image and question space by constructing synthetic dataset variants, to test how VQA methods respond to these separate shifts. 
To understand how the \emph{multiple steps and mechanisms} in recent VQA models make them robust or fragile to shifts, we compare different families (classic two-stream, transformer, and neuro-symbolic methods) by exposing them to different shifts. 
We examine multiple mechanisms to bridge domain gaps for these methods, in the challenging setting of unsupervised adaptation where no labels from the target set are available, and discuss the differences in successful versus unsuccessful attempts. 
Third, to examine the contribution of \emph{answer space differences}, we use the open-ended VQA classification formulation. Because no embedding is available for the answer options, the gap in answer spaces is more pronounced.
We compare performance across datasets and observe relations between particular modality shifts and domain robustness. 

In more detail, we compare image and question representations across nine datasets: VQA v1 and v2, VQA Abstract, Visual 7w, Visual Genome, COCO QA, CLEVR, GQA and VizWiz. We find there are large shifts in both visual and textual space, both at a low- and high-level (e.g. syntax and meaning). 
We separately apply automatic style transfer (for the visual modality) and paraphrasing (for the textual modality) to disentangle VQA methods' robustness separately to each of these artificial shifts. We also observe disparate contributions of these shifts in methods' performance across real domain gaps. 

We find evidence that neuro-symbolic, compositional models are more robust to domain shift than others, because in those methods, perception and reasoning are more disentangled. 
We argue that reasoning has the potential to be domain-independent: for example, the process of reasoning about spatial relationships can in theory be abstracted away from pixel space, thus should not need retraining if the pixel space changes. 
Inspired by the potential of perception-reasoning disentanglement, we design a two-stage domain adaptation technique to bridge domain gaps. We show that this two-stage variant is more successful than a direct, one-stage application of \cite{ganin2015unsupervised}, and a version of \cite{peng2018moment}, for recovering performance lost due to domain gaps. 

We are only aware of two prior works on domain adaptation for VQA \cite{chao2018cross,liu2020open}. 
Both of these consider supervised domain adaptation (labels present in target dataset) while we operate in an unsupervised setting (labels on source dataset only). They work with fewer datasets (2-5) and apply domain adaptation to fewer and simpler VQA methods. 
Our work can be seen as a ``reality check'' for VQA methods, similar to prior reality checks for metric learning and weakly supervised object detection \cite{Choe_2020_CVPR,musgrave2020metric}.

To summarize, our contribution is to answer the following questions:
(1) In what ways (visual, semantic, syntactic) are image-question pairs from recent VQA datasets different?
(2) What kind of dataset differences most affect VQA generalization?
(3) Which methods are more robust to synthetic visual shifts?
(4) Which methods allow more generalization when training/testing on different VQA datasets? 
(5) What domain adaptation techniques most successfully bridge domain gaps?
(6) What are the challenges of performing domain adaptation in unsupervised VQA? 

\vspace{-0.5em}
\section{Related Work}
\label{sec:related}
\vspace{-0.5em}

\noindent
\textbf{VQA method families.}
We consider three families of methods and their robustness to domain shifts.
\emph{Classic two-stream methods} \cite{anderson2018bottom,jiang2020defense,mascharka2018transparency,santoro2017simple} represent the input image and question separately, then fuse the representations to obtain an answer. Perception and cognition are entangled.
\emph{Transformer methods} \cite{chen2019uniter,gan2020villa,lu2019vilbert,tan2019lxmert,yu2020ernie} compute multiple layers of attention between entities in each modality (e.g. words to visual regions). They often use unsupervised pre-training on massive vision-language datasets (e.g. images with text captions). Other than positional encodings, these methods have no separate relational reasoning component. 
\emph{Neuro-symbolic, knowledge base, and graph methods} are conceptually distinct as they break down question-answering into modules. Some of these perform perception (e.g. recognize objects) while others perform cognition (e.g. relational reasoning about object position). 
Notable representatives include \cite{aditya2018explicit,amizadeh2020neuro,andreas2016neural,hudson2018compositional,hudson2019learning,johnson2017inferring,Mao2019NeuroSymbolic,narasimhan2018out,vedantam2019probabilistic,wang2018fvqa}.
For example, in \cite{amizadeh2020neuro,andreas2016neural,Mao2019NeuroSymbolic,vatashsky2020vqa}, entities are first parsed in a perception step, then reasoning takes a composable logic form, and questions are answered by verifying if objects satisfy a relationship implied by the question.
\cite{narasimhan2018out} extract information about objects, then look up related concepts in a knowledge base, and perform reasoning using a GCN. 
In this paper, we show that the ability to disentangle perception and reasoning enables more domain-robust question answering. 

\noindent
\textbf{Dataset bias in VQA.}
Prior work has found it is easy to introduce undesirable artifacts during dataset construction, which models can utilize to achieve misleadingly strong results. For example, \cite{Goyal_2017_CVPR} find that questions can be answered well using language priors (and bypassing the need for reasoning).
\cite{ramakrishnan2018overcoming} help a model cope with priors by discouraging it from producing an answer similar to that produced by an image-blind model. 
\cite{shah2019cycle} accomplish robustness through adversarial regularization, \cite{gokhale2020vqalol} by constructing logic compositions of existing questions, \cite{gokhale2020mutant} through semantic image mutations, and 
\cite{huang2019novel} by adding noise to the questions.
All of these are concerned with bias or lack of robustness within a single dataset, but do not examine how datasets differ in terms of image and question compositions. 

\noindent
\textbf{Domain adaptation (DA) and generalization (DG)}
cope with domain shifts, e.g. for object recognition. Unlike generalization \cite{carlucci2019domain,qiao2020learning,song2019episodic}, adaptation \cite{bousmalis2016domain,ganin2015unsupervised,hoffman2018icmlcycada,peng2018moment} assumes that some (unlabeled and/or sparsely labeled) data is available in the target domain. In the most common, classic DA setting, source and target class vocabularies overlap. 
Domain adaptation is challenging for VQA in that answer spaces do not overlap.
This setting has also been tackled in DA for object recognition, but less commonly: in partial DA \cite{cao2018partial,cao2019learning,zhang2018importance}, the target class space is a subset of the source space; and in open-set DA \cite{panareda2017open,saito2018open}, the target space could have new classes not present in the source. 
The key idea in DA is to bridge the source and target distributions and arrive at a shared representation. 
Some influential methods include gradient reversal from a domain classifier to ensure domain-agnostic features \cite{ganin2015unsupervised}, cycle-consistency \cite{hoffman2018icmlcycada}, separating shared and domain-specific features \cite{bousmalis2016domain,liu2020open}, minimizing moments of features in different domains \cite{peng2018moment}, maximizing norm which correlates with transferrability \cite{xu2019larger}, maximizing overlap between prototypes from different datasets \cite{pan2019transferrable}, etc. Methods specific to particular vision tasks also exist, e.g. for object detection where gradient reversal is applied at both the image and instance (region) level \cite{chen2018domain}, for semantic segmentation \cite{zhang2017curriculum}, etc. 
Some prior work \cite{dundar2018domain,li2018semantic,li2019bidirectional,luo2020adversarial,pizzati2020domain,rodriguez2019domain,chris_accv2018,yue2019domain} leverages style transfer techniques to bridge domain gaps, while we use style transfer and language paraphrasing to factor our shifts in the complex multi-input setting (images and question) in VQA. 

\noindent
\textbf{Prior work in domain-robust VQA.}
Our work is the first to perform fully unsupervised domain adaptation for VQA. There are only two prior works in domain-robust VQA we are aware of, but both operate in the supervised setting (i.e. some target labels are available).
\cite{chao2018cross} %
find most of the domain shift lies in questions and answers.
We consider more recent and diverse datasets, and find these contain significant image shifts as well. Further, \cite{chao2018cross} only considered a simple two-input MLP and two 2016 methods, while we consider three families of recent VQA methods. \cite{chao2018cross} is partially unsupervised; they do not use target labels to train the VQA model, but do use them to compute adaptable features.
\cite{xu-etal-2020-open} only study the shift between two datasets, and only apply domain adaptation over a non-standard method for VQA. 
In contrast to \cite{chao2018cross,xu-etal-2020-open}, we study nine datasets, and a new style transfer setting to isolate shifts in visual space.

\vspace{-0.5em}
\section{Approach}
\label{sec:approach}
\vspace{-0.5em}

We assume we have a labeled source dataset $\mathcal{D}^S = \{ \mathbf{d}^S_1, \dots, \mathbf{d}^S_i, \dots, \mathbf{d}^S_{|\mathcal{D}^S|} \}$, where each $\mathbf{d}^S_i$ is an image-question-answer triplet $\{\mathbf{v}^S_i, \mathbf{q}^S_i, a^S_i\}$. The image and question are inputs to the VQA model, and the ground-truth answer is the desired output. We also have an \emph{unlabeled} target dataset $\hat{\mathcal{D}}^T = \{\mathbf{d}^T_1, \dots, \mathbf{d}^T_j, \dots, \mathbf{d}^T_{|\hat{\mathcal{D}}^T|} \}$ where each $\mathbf{d}^T_j$ is an image-question pair $\{\mathbf{v}^T_j, \mathbf{q}^T_j\}$, and no answers are provided even in the training set. We aim to build a VQA model using $\mathcal{D}^S$ and $\mathcal{\hat{\mathcal{D}}^T}$, which can answer questions in $\hat{\mathcal{D}}^T$.
Any two datasets $\mathcal{D}^S$ and $\hat{\mathcal{D}}^T$ have potentially large domains gaps, in terms of marginal distributions (of images, questions, or answers) or conditional distributions (e.g. answers given the images or questions).
Therefore, the major challenge is to maximize the performance on $\hat{\mathcal{D}}^T$ despite the domain gaps, and our strategy is to ensure the model trained on $\mathcal{D}^S$ is as transferable to $\hat{\mathcal{D}}^T$ as possible.

We measure domain gaps for nine datasets (Sec.~\ref{sec:measure_gaps}), describe how to construct synthetic gaps to disentangle visual and linguistic shifts (Sec.~\ref{sec:synth_gaps}), and how to adapt domain adaptation techniques to bridge gaps (Sec.~\ref{sec:briding_gaps}) for individual VQA methods (Sec.~\ref{sec:vqa_models}).

\subsection{Measuring real domain gaps}
\label{sec:measure_gaps}
The first step towards building a domain-robust VQA model is to understand the multi-faceted dataset gaps.
We analyze the following datasets:
(1) VQA v1 \cite{antol2015vqa};
(2) VQA v2 \cite{Goyal_2017_CVPR};
(3) Visual Genome \cite{krishna2017visualgenome};
(4) Visual7W \cite{zhu2016visual7w};
(5) COCO-QA \cite{ren2015exploring};
(6) GQA \cite{hudson2019gqa};
(7) CLEVR \cite{johnson2017clevr};
(8) VQA Abstract \cite{antol2015vqa};
(9) VizWiz \cite{gurari2018vizwiz}.
We could measure shifts in the following distributions across datasets:
(1) $P(\mathbf{v})$;
(2) $P(\mathbf{q})$;
(3) $P(a)$;
(4) $P(\mathbf{q} | \mathbf{v})$;
(5) $P(a | \mathbf{v})$;
(6) $P(a | \mathbf{q})$;
(7) $P(a | \mathbf{v, q})$, where $\mathbf{v}$, $\mathbf{q}$ and $a$ represent image, question and answer respectively.
Here, we focus on measuring shifts in $P(\mathbf{v})$ and $P(\mathbf{q})$, 
To measure how much the corresponding distribution changes across datasets, we using Maximum Mean Discrepancy (MMD):
\begin{align}
    &\text{MMD}(\mathcal{D}^S, \hat{\mathcal{D}}^T) =  \| \mathbb{E}_{X \sim \mathcal{D}^S} [\varphi(X)] - \mathbb{E}_{Y \sim \hat{\mathcal{D}}^T} [\varphi(Y)]\|_{\mathcal{H}} \notag \\
    &= \frac{1}{n_s^2}\sum_{i=1}^{n_s}\sum_{j=1}^{n_s}k(\mathbf{x}_i, \mathbf{x}_j) + \frac{1}{n_t^2}\sum_{i=1}^{n_t}\sum_{j=1}^{n_t}k(\mathbf{y}_i, \mathbf{y}_j) \notag \\
    & - \frac{2}{n_s n_t}\sum_{i=1}^{n_s}\sum_{j=1}^{n_t}k(\mathbf{x}_i, \mathbf{y}_j)
\end{align}
where $k$ represents the RBF kernel and $n_s$, $n_t$ represent sample size in the source and target domains.
For visual representations, we use pretrained ResNet-101~\cite{he2016deep} to extract image embeddings $\{\mathbf{v}_i, \mathbf{v}_j\}$ for 10,000 randomly sampled images in each pair of datasets $\{\mathcal{D}^S, \hat{\mathcal{D}}^T\}$. We use the final 2048-D embedding as \emph{high-level semantic features}, and the spatially average-pooled embedding after conv3\_4 layer (512-D) as \emph{low-level features}.
For questions, we measure both semantic and syntactic gaps using two different representations. For the \emph{semantic representation}, we choose pre-trained BERT~\cite{devlin2018bert} to encode 10,000 randomly sampled questions $\{\mathbf{q}_i, \mathbf{q}_j\}$ from pairwise datasets $\{\mathcal{D}^S, \hat{\mathcal{D}}^T\}$. For \emph{syntactic features}, we follow the approach in \cite{gero-etal-2019-low} to extract 20 low-level features: question length, number of conjunctions, pronouns, prepositions, etc. 
We show the results in Tables \ref{tab:dataset_diffs_questions} and \ref{tab:dataset_diffs_images}.

\subsection{Constructing synthetic shifts to isolate effects}
\label{sec:synth_gaps}

As we can see in Tables \ref{tab:dataset_diffs_questions} and \ref{tab:dataset_diffs_images}, many dataset pairs differ in both their image and question distributions. Our goal is to understand precisely how different VQA methods respond to shifts in each distribution, but this is not straightforward because both modalities would affect the VQA performance. Therefore, to disentangle domain gaps arising from the image or question modality, we synthetically construct gaps in either image or question space. To do this, we use image style transfer and question paraphrasing. 

Specifically, we create stylized variants of each image in $\mathcal{D}^S$. Let $F(\mathbf{v}, \mathbf{f})$ be a style transfer function which takes in a content image $\mathbf{v}$ and style image $\mathbf{f}$ and outputs the content image now with a new style, $\mathbf{v}^{\mathbf{f}}$. 
We choose Ada-IN \cite{huang2017arbitrary} as our style transfer function $F$. We also pay extra attention to ensure colors are preserved in the style transfer process, which is important to ensure answers to color-related questions remain valid. We achieve the color preservation by converting style-transferred images into the YUV color space, and copying the UV channels from the original images.
We also experimented with the color histogram matching in \cite{Gatys_2017_CVPR}, but ultimately chose luminance-only transfer. 
We also control the transfer strength $\alpha$ in \cite{huang2017arbitrary} to avoid losing too much information. We manually verified color and answers were preserved on a small set of images.

For questions, let $G$ be a paraphrasing function,  $G(\mathbf{q}, \mathbf{g}) = \mathbf{q}^{\mathbf{g}}$, where $\mathbf{q}$ is a question and $\mathbf{g}$ is a reference ``style''. We finetuned a massively pretrained sequence-to-sequence generative T5 model \cite{raffel2019exploring} on Quora duplicate questions\footnote{https://www.kaggle.com/c/quora-question-pairs}, to shift the question $\mathbf{q}$ to a different style.

\begin{figure*}[t]
    \centering
    \includegraphics[width=\linewidth]{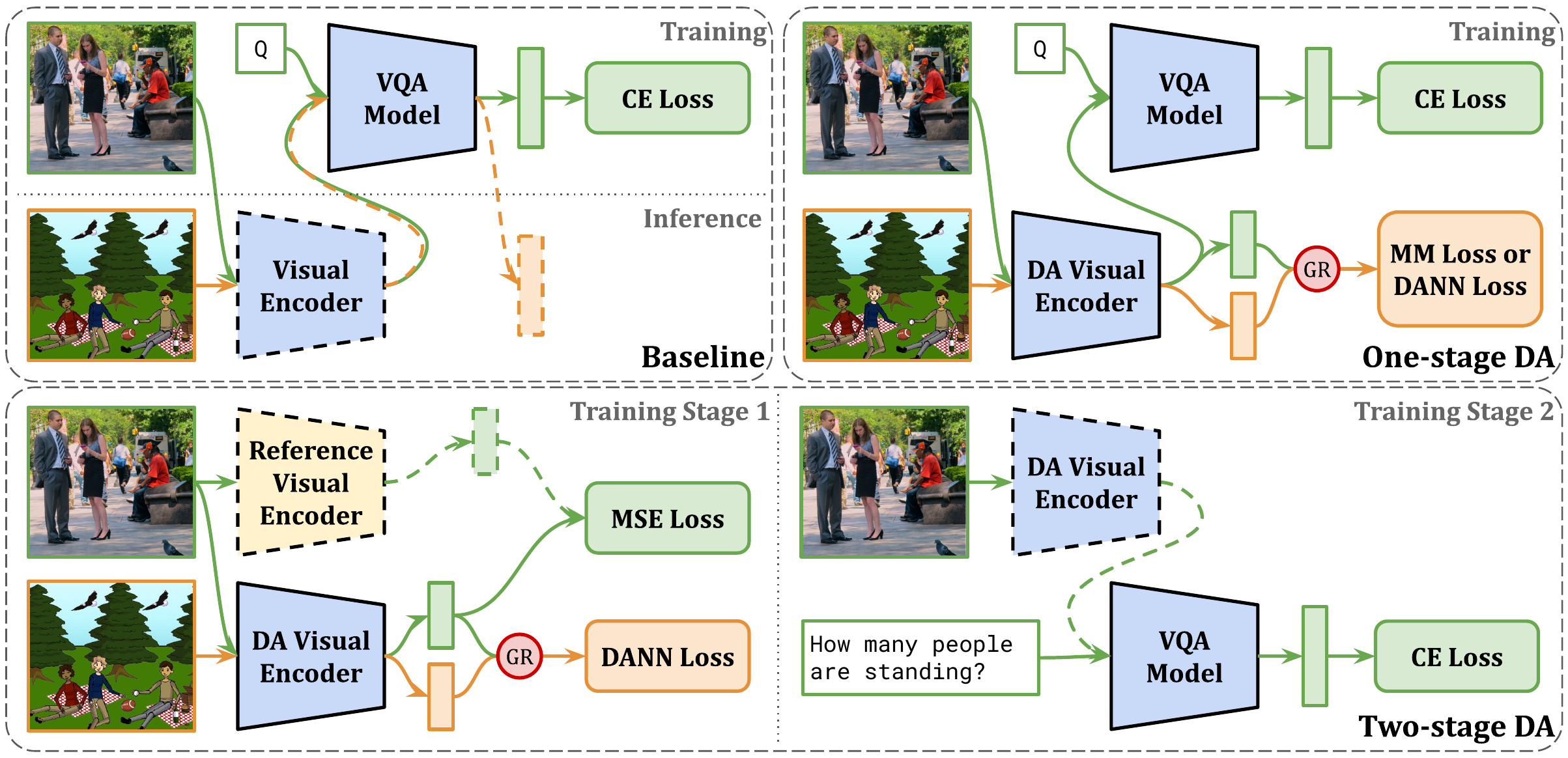}
    \caption{Illustration of the domain adaption strategies as described in Section~\ref{sec:vqa_models}. We show both training and inference stages for the baseline where no domain adaptation is applied (top left), and the training procedure for one-stage domain adaption with DANN or Moment Matching (top right). In the bottom, we show the training procedure of our proposed two-stage DANN approach. Specifically, we first train a domain-adaptive visual feature extractor in the first stage, with a MSE loss to encourage preserving semantics, and a domain confusion loss (DANN) to reduce domain gaps. Next, using extracted features from the domain-invariant extractor, we train a VQA model on source data. 
    The gradient reversal layer (GR)~\cite{ganin2015unsupervised} is only used with DANN. Dashed lines indicate no gradients due to module being frozen or for inference only.
    }
    \label{fig:synth_da}
\vspace{-1em}
\end{figure*}

\textbf{Synthetic dataset pairs}: We apply the image style transfer and question paraphrasing separately, to construct new pairs of VQA datasets that \emph{only have domain shift in one modality}. For example, by experimenting on $\mathcal{D}^S=\{\mathbf{v}, \mathbf{q}, a\}$ and $\hat{\mathcal{D}}^T=\{\mathbf{v}^\mathbf{f}, \mathbf{q}\}$, the results would reveal the model's robustness on image domain shift. If we choose $\hat{\mathcal{D}}^T=\{\mathbf{v}, \mathbf{q}^\mathbf{g}\}$, then similar experiments would show the impacts from question domain shift. Note that in both settings, the answers are kept unchanged thus the impacts from answer space shift will be eliminated.
We do not use the answers on the target domain to train, even though they are identical to those in the source domain.

\subsection{Bridging domain gaps}
\label{sec:briding_gaps}

Our goal is to ensure high accuracy on $\hat{\mathcal{D}}^T$, even though we have no ground-truth answers in the target domain as supervision. Thus, we minimize a loss of this type:
\begin{equation}
    L(\mathcal{D}^S, \hat{\mathcal{D}}^T; \theta) = L_{ce}(\mathcal{D}^S; \theta) + \lambda L_{fd}(\hat{\mathcal{D}}^T, \mathcal{D}^S; \bar{\theta})
    \label{eq:joint}
\end{equation}

In the above, $\theta$ refers to the parameters of a VQA model, to be defined in Sec.~\ref{sec:vqa_models}. $L_{ce}$ is cross-entropy loss (computed on the source dataset only), and $L_{fd}$ is a loss that computes the discrepancy between the \textbf{f}eature \textbf{d}istributions of the source and target domains, computed over images and/or questions. The bar in $\bar{\theta}$ refers to the model component over which we apply $L_{fd}$ (see Sec.~\ref{sec:vqa_models}).

For $L_{fd}$, we consider two domain adaptation strategies from object recognition, and a new variant of one of them. First, we adapt an adversarial domain classifier as described in DANN \cite{ganin2015unsupervised}, and reverse its gradient. The idea is to learn features that prevent the model $\bar{\theta}$ from being able to successfully distinguish between source and target domains. 
To successfully adapt DANN, we have to consider the differences between DA for object recognition and DA for VQA. In particular, DANN can be applied over both image and question inputs (or over intermediate representations that depend on both). 
We describe how we adapt DANN for each VQA method, in Sec.~\ref{sec:vqa_models}.
Second, we use a simplified single-source version of Moment Matching \cite{peng2018moment} which minimizes moment-related distances to reduce domain gaps.

We treat answering as 1000-way open-ended classification, and ensure the output space is the same for all datasets; we provide details in Sec.~\ref{sec:results}. 
Alternatives include answering as generation (which is challenging for automatic evaluation) or as a multiple-choice task (which may introduce biases due to the choice of the incorrect answers~\cite{chao2018cross}).

\subsection{Adaptation for VQA models}
\label{sec:vqa_models}

\noindent
\textbf{VQA models:} We analyze domain robustness of VQA models from different families:
(1) Classic two-stream methods (RelNet~\cite{santoro2017simple});
(2) Neuro-symbolic methods (NSCL~\cite{Mao2019NeuroSymbolic}); and
(3) Transformer methods (LXMERT~\cite{tan2019lxmert}).
We also test MAC~\cite{hudson2018compositional} and TbD~\cite{mascharka2018transparency}, which are hybrids of classic and neuro-symbolic methods.

\noindent
\textbf{Challenges:}
Applying domain adaptation is challenging in the unsupervised open-ended classification setting. The first challenge is the \emph{lack of labels on the target dataset, in the setting we assume.}
To the best of our knowledge, only two prior works~\cite{chao2018cross, xu-etal-2020-open} tried to tackle the domain adaptation problem for VQA. However, \emph{one leveraged multiple-choice options~\cite{chao2018cross}, and both leveraged labels in the target domains, which are not available in our setting.}
More specifically, Chao \etal \cite{chao2018cross} minimize Jensen-Shannon Divergence (JSD) to achieve domain adaptation in the multiple-choice VQA task. All datasets they investigated are derived from COCO so there is little visual domain shift, thus they only focused on dealing with question and answer/decoys shift. We noticed their improvements mostly come from minimizing JSD over answer/decoys (\ie minimizing JSD over \emph{questions} brings negligible $<0.4\%$ performance boost). In addition, \cite{chao2018cross}'s feature transformation method (impoverished VQA model without image inputs) requires labels from the target dataset. However, this is not applicable under the open-ended setting because we assume no answers and decoys for the target dataset.

A second challenge is that \emph{joint optimization of VQA with a domain adaptation objective (Eq.~\ref{eq:joint}) is unstable} because the VQA loss and DA loss may compete, making optimization difficult. This is especially true for complex, state-of-the-art VQA models. To cope with the challenge of applying domain adaptation over VQA, \cite{chao2018cross} break up adaptation and VQA training into two stages;  they primarily use a simple MLP, while we evaluate DA with recent VQA models. \cite{chao2018cross} empirically use a GAN-like approach to estimate JSD, which makes their training computationally intensive and hard to adapt to more complicated VQA models.
\cite{xu-etal-2020-open} also reports similar challenges in training a complex multitask (VQA+DA) method, and they handle it by carefully tuning the scalars for their multitask loss. Notably, they make the scalars corresponding to the unsupervised feature alignment very small (\eg, 0.003, 0.025), and the multiplier for the source classifier is also small (0.001 vs 1 for the supervised target loss). This highlights the challenge of leveraging transfer from the source domain without target labels.

\begin{table*}[t]
\begin{center}
\small
\begin{tabular}{cccccccccc}
\multicolumn{1}{l}{} & \textbf{Visual 7W} & \textbf{VG} & \textbf{VQA v1} & \textbf{VQA v2}& \textbf{COCO QA} & \textbf{CLEVR} & \textbf{VQA Abs.} & \textbf{GQA Bal.} & \textbf{VizWiz} \\
\textbf{Visual 7W}  & \cellcolor[HTML]{FFFFFF}{--} & \cellcolor[HTML]{F7FAFF}{0.04} & \cellcolor[HTML]{DAE7FD}{0.18} & \cellcolor[HTML]{DAE7FD}{0.18} & \cellcolor[HTML]{87B2F8}{0.56} & \cellcolor[HTML]{4285F4}{0.88} & \cellcolor[HTML]{DAE7FD}{0.18} & \cellcolor[HTML]{9DC0FA}{0.46} & \cellcolor[HTML]{CADDFC}{0.25} \\
\textbf{VG} & \cellcolor[HTML]{FFFBFB}{0.01} & \cellcolor[HTML]{FFFFFF}{--}    & \cellcolor[HTML]{DEEAFE}{0.16} & \cellcolor[HTML]{DEEAFE}{0.16} & \cellcolor[HTML]{8CB5F9}{0.54} & \cellcolor[HTML]{4386F5}{0.87} & \cellcolor[HTML]{DDEAFE}{0.16} & \cellcolor[HTML]{A0C2FA}{0.44} & \cellcolor[HTML]{C6DBFC}{0.27} \\
\textbf{VQA v1}       & \cellcolor[HTML]{F9DCDA}{0.06} & \cellcolor[HTML]{F8DBD8}{0.07} & \cellcolor[HTML]{FFFFFF}{--}    & \cellcolor[HTML]{FFFFFF}{0.00}    & \cellcolor[HTML]{A1C3FA}{0.44} & \cellcolor[HTML]{518FF5}{0.81} & \cellcolor[HTML]{FAFCFF}{0.03} & \cellcolor[HTML]{B6D0FB}{0.34} & \cellcolor[HTML]{C4D9FC}{0.28} \\
\textbf{VQA v2}       & \cellcolor[HTML]{F9DCD9}{0.06} & \cellcolor[HTML]{F8DAD7}{0.07} & \cellcolor[HTML]{FFFFFF}{0.00}    & \cellcolor[HTML]{FFFFFF}{--} & \cellcolor[HTML]{A1C2FA}{0.44} & \cellcolor[HTML]{508EF5}{0.81} & \cellcolor[HTML]{F9FBFF}{0.03} & \cellcolor[HTML]{B5D0FB}{0.35} & \cellcolor[HTML]{C3D9FC}{0.28} \\
\textbf{COCO QA}      & \cellcolor[HTML]{EA9088}{0.20} & \cellcolor[HTML]{EA8D85}{0.20} & \cellcolor[HTML]{EFABA5}{0.15} & \cellcolor[HTML]{EFAAA5}{0.15} & \cellcolor[HTML]{FFFFFF}{--}    & \cellcolor[HTML]{6A9FF7}{0.69} & \cellcolor[HTML]{A1C2FA}{0.44} & \cellcolor[HTML]{C8DCFC}{0.26} & \cellcolor[HTML]{83AFF8}{0.58} \\
\textbf{CLEVR}        & \cellcolor[HTML]{E8877E}{0.22} & \cellcolor[HTML]{E8857D}{0.22} & \cellcolor[HTML]{ED9F99}{0.17} & \cellcolor[HTML]{ED9F99}{0.17} & \cellcolor[HTML]{EC9790}{0.19} & \cellcolor[HTML]{FFFFFF}{--}    & \cellcolor[HTML]{528FF5}{0.81} & \cellcolor[HTML]{83AFF8}{0.58} & \cellcolor[HTML]{5D96F6}{0.76} \\
\textbf{VQA Abs.} & \cellcolor[HTML]{F9DFDD} 0.06 & \cellcolor[HTML]{F9DDDA} 0.06 & \cellcolor[HTML]{FEF5F5}{0.02} & \cellcolor[HTML]{FEF5F5}{0.02} & \cellcolor[HTML]{EFAAA4}{0.15} & \cellcolor[HTML]{EC9891}{0.19} & \cellcolor[HTML]{FFFFFF}{--}    & \cellcolor[HTML]{B7D1FB}{0.34} & \cellcolor[HTML]{C5DAFC}{0.27} \\
\textbf{GQA Bal.}          & \cellcolor[HTML]{F5C6C2}{0.10} & \cellcolor[HTML]{F4C5C1}{0.11} & \cellcolor[HTML]{F9DEDC}{0.06} & \cellcolor[HTML]{F9DEDC}{0.06} & \cellcolor[HTML]{EFABA5}{0.15} & \cellcolor[HTML]{F2BAB5}{0.13} & \cellcolor[HTML]{F8D7D4}{0.07} & \cellcolor[HTML]{FFFFFF}{--} & \cellcolor[HTML]{A4C4FA}{0.43} \\
\textbf{VizWiz}       & \cellcolor[HTML]{F9DEDC}{0.06} & \cellcolor[HTML]{FAE0DE}{0.06} & \cellcolor[HTML]{F5C8C4}{0.10} & \cellcolor[HTML]{F5C8C4}{0.10} & \cellcolor[HTML]{E67C73}{0.23} & \cellcolor[HTML]{E8847C}{0.22} & \cellcolor[HTML]{F5C8C4}{0.10} & \cellcolor[HTML]{F3BDB9}{0.12} & \cellcolor[HTML]{FFFFFF}{--}
\end{tabular}
\caption{Domain gaps in question space; red shading is MMD over 768-D BERT embeddings, blue is MMD over 20-D syntax statistics.}
\label{tab:dataset_diffs_questions}
\end{center}
\vspace{-1em}
\end{table*}

\begin{table*}[t]
\begin{center}
\small
\begin{tabular}{cccccccccc}
\multicolumn{1}{l}{} & \textbf{Visual 7W} & \textbf{VG} & \textbf{VQA v1} & \textbf{VQA v2} & \textbf{COCO QA} & \textbf{CLEVR} & \textbf{VQA Abs.} & \textbf{GQA Bal.} & \textbf{VizWiz} \\
\textbf{Visual 7W}  & \cellcolor[HTML]{FFFFFF}{--}    & \cellcolor[HTML]{FFFFFF}{0.00}    & \cellcolor[HTML]{F8FBFF}{0.01} & \cellcolor[HTML]{F6F9FF}{0.01} & \cellcolor[HTML]{F8FBFF}{0.01} & \cellcolor[HTML]{4285F4}{0.10} & \cellcolor[HTML]{6CA0F7}{0.08} & \cellcolor[HTML]{FBFCFF}{0.00} & \cellcolor[HTML]{BCD4FC}{0.04} \\
\textbf{VG} & \cellcolor[HTML]{FFFEFE}{0.01} & \cellcolor[HTML]{FFFFFF}{--} & \cellcolor[HTML]{FAFCFF}{0.00} & \cellcolor[HTML]{F6F9FF}{0.01} & \cellcolor[HTML]{F8FBFF}{0.01} & \cellcolor[HTML]{4487F5}{0.10} & \cellcolor[HTML]{6CA0F7}{0.08} & \cellcolor[HTML]{FBFCFF}{0.00} & \cellcolor[HTML]{BCD4FC}{0.04} \\
\textbf{VQA v1} & \cellcolor[HTML]{FEFAFA}{0.02} & \cellcolor[HTML]{FFFBFA}{0.02} & \cellcolor[HTML]{FFFFFF}{--} & \cellcolor[HTML]{FEFEFF}{0.00} & \cellcolor[HTML]{FEFEFF}{0.00} & \cellcolor[HTML]{4688F5}{0.10} & \cellcolor[HTML]{6CA0F7}{0.08} & \cellcolor[HTML]{F8FBFF}{0.01} & \cellcolor[HTML]{C0D6FC}{0.04} \\
\textbf{VQA v2} & \cellcolor[HTML]{FEFAF9}{0.03} & \cellcolor[HTML]{FEFAFA}{0.02} & \cellcolor[HTML]{FFFCFC}{0.01} & \cellcolor[HTML]{FFFFFF}{--} & \cellcolor[HTML]{FEFEFF}{0.00} & \cellcolor[HTML]{4A8AF5}{0.10} & \cellcolor[HTML]{6CA0F7}{0.08} & \cellcolor[HTML]{F4F8FF}{0.01} & \cellcolor[HTML]{C4D9FC}{0.03} \\
\textbf{COCO QA} & \cellcolor[HTML]{FEF6F6}{0.04} & \cellcolor[HTML]{FEF7F7}{0.04} & \cellcolor[HTML]{FEF8F8}{0.03} & \cellcolor[HTML]{FEF8F8}{0.03} & \cellcolor[HTML]{FFFFFF}{--}    & \cellcolor[HTML]{4889F5}{0.10} & \cellcolor[HTML]{6CA0F7}{0.08} & \cellcolor[HTML]{F6F9FF}{0.01} & \cellcolor[HTML]{C2D8FC}{0.03} \\
\textbf{CLEVR} & \cellcolor[HTML]{E8867E}{0.54} & \cellcolor[HTML]{E8867E}{0.54} & \cellcolor[HTML]{E9887F}{0.54} & \cellcolor[HTML]{E9877F}{0.54} & \cellcolor[HTML]{E9887F}{0.54} & \cellcolor[HTML]{FFFFFF}{--}    & \cellcolor[HTML]{4A8AF5}{0.10} & \cellcolor[HTML]{4285F4}{0.10} & \cellcolor[HTML]{629AF6}{0.09} \\
\textbf{VQA Abs.} & \cellcolor[HTML]{F0B0AA}{0.36} & \cellcolor[HTML]{F0AFAA}{0.36} & \cellcolor[HTML]{F0B0AA}{0.36} & \cellcolor[HTML]{F0AFAA}{0.36} & \cellcolor[HTML]{F0AFA9}{0.36} & \cellcolor[HTML]{E67C73}{0.59} & \cellcolor[HTML]{FFFFFF}{--}    & \cellcolor[HTML]{6CA0F7}{0.08} & \cellcolor[HTML]{669CF7}{0.08} \\
\textbf{GQA Bal.} & \cellcolor[HTML]{FEFAF9}{0.03} & \cellcolor[HTML]{FEF9F9}{0.03} & \cellcolor[HTML]{FEFAFA}{0.03} & \cellcolor[HTML]{FEF9F9}{0.03} & \cellcolor[HTML]{FEF6F6}{0.04} & \cellcolor[HTML]{E8867E}{0.54} & \cellcolor[HTML]{F0AFAA}{0.36} & \cellcolor[HTML]{FFFFFF}{--} & \cellcolor[HTML]{BBD3FC}{0.04} \\
\textbf{VizWiz} & \cellcolor[HTML]{F6CECA}{0.22} & \cellcolor[HTML]{F6CECB}{0.22} & \cellcolor[HTML]{F7D1CE}{0.21} & \cellcolor[HTML]{F7D1CE}{0.21} & \cellcolor[HTML]{F7D2CE}{0.21} & \cellcolor[HTML]{E98C84}{0.52} & \cellcolor[HTML]{EEA39C}{0.42} & \cellcolor[HTML]{F6CFCC}{0.22} & \cellcolor[HTML]{FFFFFF}{--}
\end{tabular}
\caption{Domain gaps in image space; red shading is MMD over ResNet-101 2048-D features, blue is MMD over conv3\_4 512-D features.}
\label{tab:dataset_diffs_images}
\end{center}
\vspace{-1em}
\end{table*}

\noindent
\textbf{Baseline and one-stage approaches:}
We report the performance of two reference models: (1) the accuracy on the source dataset, which indicates model capacity, and (2) the accuracy on the target dataset assuming target labels are fully available, which serves as an empirical upper bound for domain adaptation. 
The simplest baseline is directly applying a model trained on a source dataset, on test data from the target domain, without any domain adaptation. The training and inference procedure is illustrated in Fig.~\ref{fig:synth_da}.
As another baseline, we also investigated an end-to-end pipeline to combine the DANN training with VQA training, shown as ``One-stage DA''. Specifically, we added the domain discrimination loss and reversed its gradients to update the visual representations.
However, it is non-trivial to find the best place for applying domain discrimination for different VQA methods. For example, for MAC we added a linear classifier to distinguish the domains and applied the DANN loss on the visual embedding before feeding them into the MAC unit. 
In addition to DANN, we experimented with moment matching~\cite{peng2018moment} where the first- and second-order moments are enforced to align across domains. In this case the gradient reversal layer is no longer needed.

\noindent
\textbf{Proposed two-stage DA approach:} 
To better cope with the challenges, we also propose a two-stage approach to build a domain-invariant feature extractor and VQA module sequentially. A figurative illustration of the process is shown in Fig.~\ref{fig:synth_da} (bottom). The motivation for breaking up domain adaptation and VQA modeling is to stabilize the training for greater robustness.
The idea is partially inspired by neuro-symbolic methods, which separate perception (in this case, feature extraction) and reasoning (the VQA model after feature extraction).
Our two-stage strategy is summarized as:
\begin{enumerate}[nolistsep,noitemsep,leftmargin=*]
    \item Extract features for images in the source dataset, as defined in the VQA method (\eg use pre-trained ResNet).
    \item Train a domain-invariant feature extractor with both source and target datasets (without labels), using (a) an MSE loss which encourages the extracted features on source dataset to preserve semantics, and (b) a BCE loss with gradient reversal layer to prevent distinguishing the source and target domains. 
    \item Apply the backbone from step 2 to extract visual features and train a VQA model on the source dataset. 
    \item Take the visual feature extractor from step 2 and VQA model from step 3, then feed in the target dataset and evaluate the performance.
\end{enumerate}

\noindent
\textbf{VQA method specifics:}
Each VQA method extracts features in a particular way, resulting in small variances in our two-stage DA implementation. For MAC and TbD, feature extraction is executed with ResNet-101 prior to training the VQA model, following the methodology outlined previously.
NSCL uses ResNet-34 to extract features from different regions in the image (region proposals via a pre-trained Mask R-CNN \cite{He_2017_ICCV}), and allows for NSCL to fine-tune ResNet-34 during training. To most closely follow our two-stage methodology, we replace the pretrained ResNet-34 with a frozen ResNet-34 backbone trained in step 2.
RelNet uses 4 convolutional layers to extract features from the images. We used a pre-trained set of these convolutional layers to export the source features for VQA and DA.
For LXMERT, the initial visual features are from pre-trained Faster R-CNN and processed by vision-only transformer layers. We kept the Faster R-CNN backbone untouched and fine-tuned the transformer layers to be domain-invariant. 

\vspace{-0.5em}
\section{Experimental Validation}
\label{sec:results}
\vspace{-0.5em}

We show four groups of results: shifts in image and question space for nine datasets (Sec.~\ref{sec:dataset_shifts}), robustness of five methods to synthetic shifts in visual or textual space using the CLEVR dataset (Sec.~\ref{sec:results_synth}), different ways to apply unsupervised domain adaptation using MAC on three datasets (Sec.~\ref{sec:results_da}), and finally robustness of two methods using eight real dataset pairs (Sec.~\ref{sec:results_real}).

\begin{table*}[t]
    \centering
    \setlength{\tabcolsep}{0.5em}
    \renewcommand\tabularxcolumn[1]{m{#1}}
    \begin{tabularx}{\linewidth}{
    | >{\hsize=1.05\hsize}>{\raggedright\arraybackslash}X
    || >{\hsize=0.9\hsize}>{\centering\arraybackslash}X
    || >{\hsize=0.9\hsize}>{\centering\arraybackslash}X
    | >{\hsize=1.15\hsize}>{\centering\arraybackslash}X
    || >{\hsize=1.05\hsize}>{\centering\arraybackslash}X
    | >{\hsize=1.05\hsize}>{\centering\arraybackslash}X
    | >{\hsize=0.9\hsize}>{\centering\arraybackslash}X |
    }
    \hline
    Method / Type &	Source Acc. & Target Acc. (direct) & Target Acc. (2-stage DANN) & Target Acc. (10\% scratch) & Target Acc. (10\% finetune) & Target Acc. (full) \\
    \hline
    NSCL (NS) & \textbf{98.0} & \textbf{59.7} & \textbf{68.6} & 60.0 & 75.8 & \textbf{95.9} \\
    \hline
    MAC (NS/CL) & 93.4 & \textbf{62.6} & \textbf{65.2} & \textbf{84.6} & \textbf{82.1} & 88.6 \\
    \hline
    TbD (NS/CL) & \textbf{99.1} & 36.3 & 41.3 & \textbf{72.5} & \textbf{84.2} & \textbf{95.3} \\
    \hline
    RelNet (CL) & 93.7 & 44.8 & 47.2 & 61.5 & 77.1 & 91.4 \\
    \hline
    LXMERT (TR) & 94.8 & 58.0 & -- & 60.9 & 65.9 & 91.3 \\
    \hline
    \end{tabularx}
    \caption{Method robustness on CLEVR, using style transfer of the original images (domain shift in image space). We bold the best two results per column. The most important columns are Target (direct) and Target (2-stage DANN) as they require no supervision on the target. We observe neuro-symbolic methods are most robust. -- means performance degraded on LXMERT with DANN.}
    \label{tab:synth_clevr}
\vspace{-1em}
\end{table*}
\begin{table}[t]
    \centering
    \small
    \setlength{\tabcolsep}{0.5em}
    \begin{tabularx}{\linewidth}{
    | >{\hsize=2.0\hsize}>{\centering\arraybackslash}X 
    || >{\hsize=0.8\hsize}>{\centering\arraybackslash}X 
    || >{\hsize=0.8\hsize}>{\centering\arraybackslash}X
    | >{\hsize=0.8\hsize}>{\centering\arraybackslash}X
    || >{\hsize=0.8\hsize}>{\centering\arraybackslash}X
    | >{\hsize=0.8\hsize}>{\centering\arraybackslash}X |
    }
    \hline
    Methods & Q & I1    & I1+Q  & I2    & I2+Q  \\ \hline
    NSCL (NS) & -- & \textbf{71.0} & -- & 60.6 & -- \\ \hline
    MAC (NS/CL) & 52.2 & 45.9 & 28.1 & 60.9 & 37.9 \\ \hline
    TbD (NS/CL) & \textbf{52.9} & 55.7 & \textbf{36.1} & \textbf{70.4} & \textbf{42.6} \\ \hline
    RelNet (CL) & 49.6 & 20.5 & 19.1 & 46.2 & 31.6 \\ \hline
    LXMERT (TR) & \textbf{53.4} & 50.6 & \textbf{36.6} & 58.0 & 40.5 \\ \hline
    \end{tabularx}
    \caption{Method robustness on CLEVR. We show performance under artificial \emph{Question} shifts, followed by \emph{Image} shifts with two styles (resulting in I1 and I2), and two settings where both Image and Question shifts are applied (I+Q). -- means we were unable to test on NSCL since their semantic parser is not open-sourced. We bold the best result and those within 1\% of the best.}
    \label{tab:synth_clevr_both}
\vspace{-1em}
\end{table}

\subsection{Domain shifts in nine datasets}
\label{sec:dataset_shifts}

Tables \ref{tab:dataset_diffs_questions} and \ref{tab:dataset_diffs_images} show how the questions and images in nine datasets differ. Each table is a composition of two triangles. In Table  \ref{tab:dataset_diffs_questions}, the lower triangle contains Maximum Mean Discrepancy (MMD) statistics using BERT embeddings, while the upper triangle shows MMD statistics using syntax features (Sec.~\ref{sec:measure_gaps}). MMD computes how different two distributions are, with higher values indicating larger difference. The shading ranges from white to red/blue, with darker, more vivid colors indicating larger values. 

In the lower triangle of Table \ref{tab:dataset_diffs_questions}, we observe that Visual 7W and Visual Genome (VG) are similar, and VQA v1 and v2 are similar, as expected. 
GQA is similar to VQA v1/v2 in terms of semantics (captured through BERT), but it is different in terms of syntax.
VQA Abstract is much more similar to VQA v1/v2 in terms of syntax than to other datasets (blue triangle), but in terms of semantic content (red triangle), it is also fairly similar to Visual 7W and VG. COCO QA and CLEVR stand out from other datasets both in terms of semantics and syntax (both rows/columns for COCO QA and CLEVR have high values except on diagonal), but CLEVR's syntax (darker blue) stands out more than COCO QA's syntax (lighter blue), while in terms of semantics they are similarly unique. GQA and VizWiz are also relatively unique, but less so than CLEVR. In Sec.~\ref{sec:results_real}, we show how these shifts affect cross-dataset performance.

Some dataset pairs that were distinct in terms of questions are similar in terms of images, and vice versa, as shown in Table \ref{tab:dataset_diffs_images}. COCO QA is now fairly similar to other datasets (in terms of images), but VQA Abstract and VizWiz become more unique (darker shading) than in Table \ref{tab:dataset_diffs_questions}; they are two of the three rows/columns with high values, in addition to CLEVR. Results are generally consistent in the lower/upper triangles (from ResNet layers closer to the output or input, respectively) except that in higher dimensions (lower triangle), absolute MMD scores are larger.

\subsection{Methods' robustness to synthetic domain shifts}
\label{sec:results_synth}

Tables \ref{tab:synth_clevr} and \ref{tab:synth_clevr_both} show how robust different VQA methods are to synthetic shifts on the CLEVR dataset. In Table \ref{tab:synth_clevr}, we show robustness to visual shifts. We evaluate performance by the method on the original CLEVR dataset, performance of the model trained on CLEVR and applied in the shifted setting (\eg style-transferred images) directly, target performance with unsupervised domain adaptation (specifically, 2-stage DANN), and three supervised settings for comparison -- two that use 10\% of the target training data, and one that uses 100\% of the target training data. 
We use the default recommended hyperparameters without exhaustive search.
We observe all methods' performance drops in the Target setting compared to Source, as expected. However, in Target (direct) and Target (2-stage DANN), both of which do not use labels on the target, NSCL and MAC (both neuro-symbolic or NS hybrid) retain the best performance. Using a small amount of target data for fine-tuning, MAC and TbD (both NS hybrids) perform best.

Table \ref{tab:synth_clevr_both} demonstrates each method's change in performance when evaluated with paraphrased questions (first column), style-transferred images using two separate styles, I1 and I2 (second and fourth columns), and combined question and image shifts (third and fifth columns). 
LXMERT is most robust to question shifts, likely due to its extensive pre-training on language data, followed by TbD. Neuro-symbolic or hybrid methods (NSCL or TbD) are most robust to image shifts, consistent with our hypothesis.

\subsection{Domain adaptation for synthetic shifts}
\label{sec:results_da}

\begin{table}[t]
    \centering
    \small
    \setlength{\tabcolsep}{0.5em}
    \begin{tabularx}{\linewidth} {
      | >{\hsize=1.85\hsize}>{\raggedright\arraybackslash}X 
      || >{\hsize=0.7\hsize}>{\centering\arraybackslash}X 
      | >{\hsize=0.7\hsize}>{\centering\arraybackslash}X 
      | >{\hsize=0.75\hsize}>{\centering\arraybackslash}X | }
    \hline
        & VQA v2    & CLEVR & GQA Bal. \\
    \hline
    Source Accuracy & 54.0 & 95.8 & 44.6 \\ \hline
    \hline
    Target (direct) & 41.0 & 45.9 & 37.3 \\
    \hline
    Target (1-stage DANN) & 42.2 & 45.7 & 37.4 \\
    \hline
    Target (1-stage MM) & 42.6 & 46.6 & \textbf{38.6} \\
    \hline
    Target (2-stage DANN) & \textbf{42.8} & \textbf{46.7} & 38.5 \\ \hline
    \hline
    Target (full) & 49.1 & 90.0 & 42.1 \\
    \hline
    \end{tabularx}
    \caption{Different DA methods on MAC (NS/CL), image shift.} 
    \label{tab:synth_da}
\vspace{-1em}
\end{table}

In Table~\ref{tab:synth_da}, we evaluate different domain adaptation strategies with MAC as the backbone model on VQA v2, CLEVR, and GQA Balanced, where artificial domain shifts are created in the image space.
By comparing Source and Target (full) accuracy, we deduce the image style transfer preserves the information required for VQA as accuracy only drops slightly. However, in all datasets, we see quite significant performance drop if a trained model is directly applied to the corresponding target dataset. The domain adaptation strategies (1-stage DANN \cite{ganin2015unsupervised} and Moment Matching \cite{peng2018moment}, and our 2-stage DANN) help to different degree. 
Our proposed 2-stage DANN is always significantly better than then 1-stage DANN, and better than the 1-stage MM on two of three datasets. Note that differences between methods are significant in that the range between Target (direct) and Target (full) is very small for two of the three datasets. It is worth mentioning that training the 1-stage DANN baseline is highly unstable as the optimization is more difficult. 
We repeated the experiments multiple times and only preserved the 1-stage DANN models that did not collapse. %
Because of the challenges mentioned, on real dataset shifts, we only achieved marginal gains using domain adaptation, over directly applying the source model, consistent with prior work \cite{chao2018cross,xu-etal-2020-open}.

\subsection{Generalization under real domain shifts}
\label{sec:results_real}
\begin{table}[t]
    \centering
    \small
    \setlength{\tabcolsep}{0em}
    \begin{tabularx}{\linewidth}{
    | >{\hsize=0.4\hsize}>{\centering\arraybackslash}X 
    | >{\hsize=1.3\hsize}>{\centering\arraybackslash}X 
    | >{\hsize=1.3\hsize}>{\centering\arraybackslash}X 
    || >{\hsize=\hsize}>{\centering\arraybackslash}X 
    | >{\hsize=\hsize}>{\centering\arraybackslash}X 
    || >{\hsize=\hsize}>{\centering\arraybackslash}X
    | >{\hsize=\hsize}>{\centering\arraybackslash}X |
    }
    \hline
    \multirow{2}{*}{} & \multicolumn{2}{c||}{Datasets} & \multicolumn{4}{c|}{Accuracy (\%)} \\
    \cline{2-7}
	& $\mathcal{A}$ & $\mathcal{B}$ & $\mathcal{A}$ & $\mathcal{B}$ & $\mathcal{A} \rightarrow \mathcal{B}$ & $\mathcal{B} \rightarrow \mathcal{A}$ \\
    \hline
    \multirow{4}{*}{\rotatebox[origin=c]{90}{MAC}} & \multirow{4}{*}{VQA v2} & CLEVR &
    \multirow{4}{*}{53.3} & 95.9 & \cellcolor{blue!15.6}29.8 & \cellcolor{red!9.7}18.7 \\ \cline{3-3}\cline{5-7}
    & &	GQA Bal. & & 44.4 & \cellcolor{blue!36.0}32.0 & \cellcolor{red!40.1}35.6 \\ \cline{3-3}\cline{5-7}
    & &	VQA Abs. & & 48.3 & \cellcolor{blue!34.8}33.6 & \cellcolor{red!32.8}31.7 \\ \cline{3-3}\cline{5-7}
    & &	VG & & 33.3 & \cellcolor{blue!39.3}26.2 & \cellcolor{red!34.7}23.1 \\ \hline
    \multirow{4}{*}{\rotatebox[origin=c]{90}{LXMERT}} & \multirow{4}{*}{VQA v2} & CLEVR &
    \multirow{4}{*}{67.6} & 84.9 & \cellcolor{blue!18.6}31.6 & \cellcolor{red!20.5}34.8 \\ \cline{3-3}\cline{5-7}
    & &	GQA Bal. & & 58.2 & \cellcolor{blue!43.4}50.5 & \cellcolor{red!44.2}51.5 \\ \cline{3-3}\cline{5-7}
    & &	VQA Abs. & & 56.3 & \cellcolor{blue!30.5}34.3 & \cellcolor{red!30.7}34.6 \\ \cline{3-3}\cline{5-7}
    & &	VG & & 41.0 & \cellcolor{blue!44.8}36.7 & \cellcolor{red!38.3}31.4 \\ \hline
    \end{tabularx}
    \caption{Robustness across VQA datasets; best viewed in color.}
    \label{tab:cross_dataset}
\end{table}

\begin{figure}[t]
    \centering
    \begin{minipage}[c]{.3\linewidth}
        \centering
        \includegraphics[width=\linewidth]{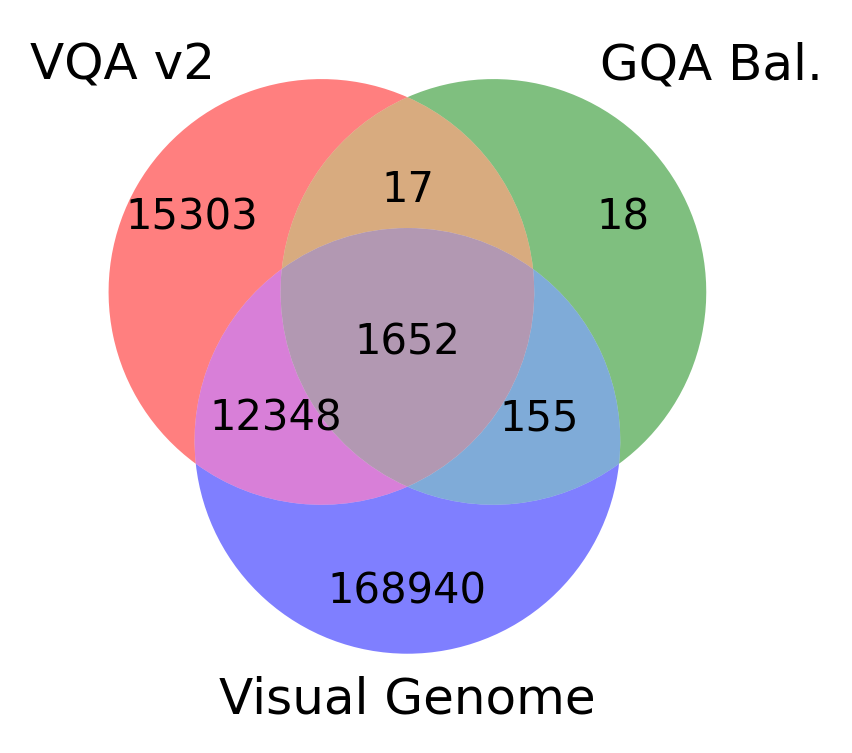}
    \end{minipage}%
    \begin{minipage}[c]{0.6\linewidth}
        \centering
        \footnotesize
        \setlength{\tabcolsep}{0em}
        \begin{tabularx}{\linewidth}{
        | >{\hsize=0.55\hsize}>{\centering\arraybackslash}X
        | >{\hsize=1.45\hsize}>{\centering\arraybackslash}X |
        }
        \hline
        Dataset & Most Frequent Answers \\ \hline
        VQA v2  & yes, no, 2, 1, white, 3, 4, \dots  \\
        CLEVR   & no, yes, 1, 0, small, rubber, \dots  \\
        GQA Bal. & no, yes, left, right, man, \dots \\
        VQA Abs. & yes, no, 2, 1, red, 3, white, \dots \\
        VG & 1, white, 2, daytime, black, \dots \\
        \hline
        \end{tabularx}
    \end{minipage}
    \caption{Venn diagram of answer vocabulary of three datasets. A large portion of answers are not shared across datasets, and the distribution (\eg most frequent answers) may differ as well.}
    \label{tab:answer_space}
\end{figure}

Table \ref{tab:cross_dataset} shows the robustness of two recent VQA methods among five datasets: VQA v2, CLEVR, GQA Balanced, VQA Abstract and Visual Genome.
These datasets have different answer spaces, as shown in Fig.~\ref{tab:answer_space}. Since the final classification layer is coupled with the answer vocabulary, models trained on one dataset cannot be directly applied to another.
To mitigate this issue, we obtain a shared 1000-class answer space by computing the 1000 most common answers across all five selected datasets.
We report training and evaluating a model on the same dataset (\ie Acc of $\mathcal{A}$ and $\mathcal{B}$), and training on one and evaluating on the other (\eg Acc of $\mathcal{A} \rightarrow \mathcal{B}$ denotes training on $\mathcal{A}$ and evaluating on $\mathcal{B}$). The accuracy is calculated on the validation split for individual datasets (except for GQA where we use testdev split as recommended), and is obtained by matching the top-1 prediction with the ground-truth answer(s). %

Since source/target datasets have different upper bounds (\ie $\mathcal{B}$ Acc), we normalize the transferred accuracy by dividing by $\mathcal{B}$, and illustrate the relative normalized performance using the intensity of shading: darker background of a cell indicates higher ratio of the transferred accuracy and the source/target accuracy. Blue backgrounds measure how well a transferred model $\mathcal{A} \rightarrow \mathcal{B}$ performs compared to its upper bound, as they are all transferring from the same source $\mathcal{A}$, while red backgrounds measure how well different source models $\mathcal{B} \rightarrow \mathcal{A}$ transfer to the same target $\mathcal{A}$.

\begin{table}[t]
    \centering
    \small
    \setlength{\tabcolsep}{0em}
    \begin{tabularx}{\linewidth}{
    | >{\hsize=1\hsize}>{\centering\arraybackslash}X
    || >{\hsize=\hsize}>{\centering\arraybackslash}X
    | >{\hsize=\hsize}>{\centering\arraybackslash}X
    | >{\hsize=\hsize}>{\centering\arraybackslash}X
    | >{\hsize=\hsize}>{\centering\arraybackslash}X |
    }
    \hline
    \multirow{2}{*}{Datasets} & \multicolumn{2}{c|}{Image} & \multicolumn{2}{c|}{Question} \\ \cline{2-5}
    & Appearance & Semantic & Syntactic & Semantic \\
    \hline
    CLEVR       & High      & High      & High      & High \\ \hline
    GQA Bal.    & Low       & Low       & Med. High & Medium \\ \hline
    VQA Abs.    & Med. High & Med. High & Low       & Low \\ \hline
    VG          & Low       & Low       & Med. Low  & Medium \\ \hline
    \end{tabularx}
    \caption{Summary of shifts, VQA-v2 $\leftrightarrow$ selected datasets.}
    \label{tab:selected_datasets}
\vspace{-1em}
\end{table}

By comparing the accuracy on the training and evaluation datasets, we see that in most cases LXMERT (TR) generalizes better across datasets than MAC (NS/CL). We hypothesize that transformer-based methods like LXMERT benefit from their massive pre-training (which includes disjoint GQA and VQA v2 data).
We also observe that GQA and Visual Genome are more useful sources when transferring knowledge to VQA v2, compared to CLEVR. This observation is consistent with our statistical analysis in Tables~\ref{tab:dataset_diffs_questions} and \ref{tab:dataset_diffs_images}, and for simplicity we extracted relevant information in Table~\ref{tab:selected_datasets}. 
We see that GQA Balanced and Visual Genome are similar to VQA v2 in multiple aspects. We also note that GQA Balanced has \emph{smaller semantic shifts than syntactic shifts} with respect to VQA v2, while VG has smaller syntactic than semantic shifts with VQA v2. This makes GQA Balanced more helpful as a source dataset (darker shading for GQA$\rightarrow$VQA-v2 than for VG$\rightarrow$VQA-v2 in Table \ref{tab:cross_dataset}, for both MAC and LXMERT). 
Finally, the only case MAC is more robust than LXMERT (in terms of shading) is VQA-v2$\leftrightarrow$VQA-Abstract, which is the dataset with largest visual shifts after CLEVR. One possibility is that LXMERT is better suited to deal with question shifts and MAC with visual shifts, because of its neuro-symbolic nature and dedicated perception module.

\vspace{-0.5em}
\section{Conclusion}
\label{sec:conclusion}
\vspace{-0.5em}

We showed domain differences between VQA datasets can come from the visual and linguistic space; different methods are more susceptible to visual or linguistic shifts, and high-level semantic shifts make methods more fragile than syntactic ones. 
We found neuro-symbolic methods are more robust to synthetic visual-only domain shifts and some real dataset shifts, but transformer methods handle real linguistic and some visual shifts better due to pretraining.
We demonstrated that while unsupervised domain adaptation in VQA is challenging, better gains can be made through a two-stage DANN which shares similar intuition as neuro-symbolic methods.
In the future, we will explicitly handle shifts in answer space, and develop DA techniques that can flexibly choose how much to adapt over each modality,

\vspace{1em}
\noindent \textbf{Acknowledgements:} This material is based upon work supported by the National Science Foundation under Grant No. 1718262. It was also supported by the Univ. of Pittsburgh Momentum Fund, and Google/Amazon/Adobe gifts. 
We thank the reviewers and AC for their suggestions.
\clearpage

{\small
\bibliographystyle{ieee_fullname}
\bibliography{refs}
}

\end{document}